\documentclass{article}

\usepackage[nonatbib,final]{neurips_2020}

\usepackage[utf8]{inputenc} 
\usepackage[T1]{fontenc}    
\usepackage[dvipsnames]{xcolor}
\usepackage[pagebackref=true,breaklinks=true,letterpaper=true,colorlinks,bookmarks=false, citecolor=ForestGreen]{hyperref}     
\usepackage{url}            
\usepackage{booktabs}       
\usepackage{amsfonts}       
\usepackage{nicefrac}       
\usepackage{microtype}      

\usepackage{graphicx}
\usepackage{comment}
\usepackage{amsmath,amssymb} 
\usepackage{xspace}
\usepackage{caption}
\usepackage{mathtools}
\usepackage{graphicx}
\usepackage{soul}
\usepackage{xspace}
\usepackage[subrefformat=parens]{subfig}
\usepackage{cleveref}
\usepackage{enumitem}
\usepackage[square,numbers,sort&compress]{natbib}

\usepackage{bbding}
\usepackage{pifont}
\usepackage{wasysym}
\usepackage{amssymb}
\usepackage{float}
\usepackage{dblfloatfix}
\usepackage{footnote}
\makesavenoteenv{tabular}
\makesavenoteenv{table}
\makesavenoteenv{figure}
\usepackage{xspace}
\usepackage{multirow}

\def\onedot{.}

\def\ie{\emph{i.e}\onedot} 
 
 \def\vs{\emph{vs}\onedot}
 
\def\etal{\emph{et al}\onedot}
\crefname{section}{\S}{\S\S}
\crefname{subsection}{\S}{\S\S}
\crefformat{table}{Table~#2#1#3}
\crefformat{figure}{Fig.~#2#1#3}
\crefformat{equation}{Eqn~#2#1#3}

\def\OURS{SwAV\xspace}
\def\OURSFULL{Swapping Assignments between Views\xspace}
\newcommand{\appendixName}{Appendix}

\newcommand{\resnetfifty}{R50}

\newcommand{\ImNet}{ImageNet\xspace}

\newcommand{\iNat}{iNat18\xspace}
\newcommand{\VOCseven}{VOC07\xspace}
\newcommand{\VOCseventwelve}{VOC07+12\xspace}
\newcommand{\Places}{Places205\xspace}

\newcommand{\code}[1]{\texttt{#1}}
\newcommand{\com}[1]{{\color{CadetBlue}\texttt{#1}}}

\title{
Unsupervised Learning of Visual Features \\
by Contrasting Cluster Assignments
}

\author{Mathilde Caron$^{1,2}$ \And Ishan Misra$^2$ \And \hspace{0.2cm} Julien Mairal$^1$ \hspace{-0.2cm} \AND \hspace{0.5cm} Priya Goyal$^2$ \And \hspace*{0.3cm} Piotr Bojanowski$^2$ \And Armand Joulin$^2$ \AND \normalfont $^1$ Inria$^*$ \And \normalfont $^2$ Facebook AI Research}

\begin{document}

\maketitle

\begin{abstract}
Unsupervised image representations have significantly reduced the gap with supervised pretraining, notably with the recent achievements of contrastive learning methods.
These contrastive methods typically work online and rely on a large number of explicit pairwise feature comparisons, which is computationally challenging.
In this paper, we propose an online algorithm, \OURS, that takes advantage of contrastive methods without requiring to compute pairwise comparisons.
Specifically, our method simultaneously clusters the data while enforcing consistency between cluster assignments produced for different augmentations (or ``views'') of the same image, instead of comparing features directly as in contrastive learning.
Simply put, we use a ``swapped'' prediction mechanism where we predict the code of a view from the representation of another view.
Our method can be trained with large and small batches and can scale to unlimited amounts of data.
Compared to previous contrastive methods, our method is more memory efficient since it does not require a large memory bank or a special momentum network.
In addition, we also propose a new data augmentation strategy, \texttt{multi-crop}, that uses a mix of views with different resolutions in place of two full-resolution views, without increasing the memory or compute requirements.
We validate our findings by achieving $75.3\%$ top-1 accuracy on ImageNet with ResNet-50, as well as surpassing supervised pretraining on all the considered transfer tasks.

\end{abstract}

\newcommand\blfootnote[1]{%
  \begingroup
  \renewcommand\thefootnote{}\footnotetext{#1}%
  \addtocounter{footnote}{-1}%
  \endgroup
}
\blfootnote{* Univ. Grenoble Alpes, Inria, CNRS, Grenoble INP, LJK, 38000 Grenoble, France}
\blfootnote{Correspondence to \texttt{mathilde@fb.com}}
\blfootnote{Code: \url{https://github.com/facebookresearch/swav}}


\section{Introduction}

Unsupervised visual representation learning, or self-supervised learning, aims at obtaining features without using manual annotations and is rapidly closing the performance gap with supervised pretraining in computer vision~\cite{chen2020simple,he2019momentum,misra2019self}.
Many recent state-of-the-art methods build upon the instance discrimination task that considers each image of the dataset (or ``instance'') and its transformations as a separate class~\cite{dosovitskiy2016discriminative}. 
This task yields representations that are able to discriminate between different images, while achieving some invariance to image transformations. 
Recent self-supervised methods that use instance discrimination rely on a combination of two elements: (i) a contrastive loss~\cite{hadsell2006dimensionality} and (ii) a set of image transformations. 
The contrastive loss removes the notion of instance classes by directly comparing image features while the image transformations define the invariances encoded in the features.
Both elements are essential to the quality of the resulting networks~\cite{misra2019self,chen2020simple} and our work improves upon both the objective function and the transformations.

The contrastive loss explicitly compares pairs of image representations to push away representations from different images while pulling together those from transformations, or views, of the same image.
Since computing all the pairwise comparisons on a large dataset is not practical, most implementations approximate the loss by reducing the number of comparisons to random subsets of images during training~\cite{chen2020simple,he2019momentum,wu2018unsupervised}.
An alternative to approximate the loss is to approximate the task---that is to relax the instance discrimination problem.
For example, clustering-based methods discriminate between groups of images with similar features instead of individual images~\cite{caron2018deep}. 
The objective in clustering is tractable, but it does not scale well with the dataset as it requires a pass over the entire dataset to form image ``codes'' (\ie, cluster assignments) that are used as targets during training.
In this work, we use a different paradigm and propose to compute the codes online while enforcing consistency between codes obtained from views of the same image.
Comparing cluster assignments allows to contrast different image views while not relying on explicit pairwise feature comparisons.
Specifically, we propose a simple ``swapped'' prediction problem where we predict the code of a view from the representation of another view.
We learn features by \textbf{Sw}apping \textbf{A}ssignments between multiple \textbf{V}iews of the same image (\textbf{\OURS}).
The features and the codes are learned online, allowing our method to scale to potentially unlimited amounts of data.
In addition, \OURS works with small and large batch sizes and does not need a large memory bank~\cite{wu2018unsupervised} or a momentum encoder~\cite{he2019momentum}.

Besides our online clustering-based method, we also propose an improvement to the image transformations.
Most contrastive methods compare one pair of transformations per image, even though there is evidence that comparing more views during training improves the resulting model~\cite{misra2019self}. 
In this work, we propose \texttt{multi-crop} that uses smaller-sized images to increase the number of views while not increasing the memory or computational requirements during training. 
We also observe that mapping small parts of a scene to more global views significantly boosts the performance.
Directly working with downsized images introduces a bias in the features~\cite{touvron2019fixing}, which can be avoided by using a mix of different sizes. 
Our strategy is simple, yet effective, and can be applied to many self-supervised methods with consistent gain in performance.

We validate our contributions by evaluating our method on several standard self-supervised benchmarks.
In particular, on the ImageNet linear evaluation protocol, we reach $75.3\%$ top-$1$ accuracy with a standard ResNet-50, and $78.5\%$ with a wider model.
We also show that our \texttt{multi-crop} strategy is general, and improves the performance of different self-supervised methods, namely SimCLR~\cite{chen2020simple}, DeepCluster~\cite{caron2018deep}, and SeLa~\cite{asano2019self}, between $2\%$ and $4\%$ top-1 accuracy on ImageNet.
Overall, we make the following contributions:
\begin{itemize}[leftmargin=*]
\itemsep0.05em 

\item
We propose a scalable online clustering loss that improves performance by $+2\%$ on ImageNet and works in both large and small batch settings without a large memory bank or a momentum encoder.
\item
We introduce the \texttt{multi-crop} strategy to increase the number of views of an image with no computational or memory overhead.
We observe a consistent improvement of between $2\%$ and $4\%$ on \ImNet with this strategy on several self-supervised methods.
\item
Combining both technical contributions into a single model, we improve the performance of self-supervised by $+4.2\%$ on ImageNet with a standard ResNet and outperforms supervised ImageNet pretraining on multiple downstream tasks.
This is the first method to do so without finetuning the features, \ie, only with a linear classifier on top of frozen features.
\end{itemize}


\section{Related Work}
\label{sec:related}

\paragraph{Instance and contrastive learning.}
Instance-level classification considers each image in a dataset as its own class~\cite{bojanowski2017unsupervised,dosovitskiy2016discriminative,wu2018unsupervised}.
Dosovitskiy~\etal~\cite{dosovitskiy2016discriminative} assign a class explicitly to each image and learn a linear classifier with as many classes as images in the dataset.
As this approach becomes quickly intractable, Wu~\etal~\cite{wu2018unsupervised} mitigate this issue by replacing the classifier with a memory bank that stores previously-computed representations.
They rely on noise contrastive estimation~\cite{gutmann2010noise} to compare instances, which is a special form of contrastive learning~\cite{hjelm2018learning,oord2018representation}.
He~\etal~\cite{he2019momentum} improve the training of contrastive methods by storing representations from a momentum encoder instead of the trained network.
More recently, Chen~\etal~\cite{chen2020simple} show that the memory bank can be entirely replaced with the elements from the same batch if the batch is large enough.
In contrast to this line of works, we avoid comparing every pair of images by mapping the image features to a set of trainable prototype vectors.

\paragraph{Clustering for deep representation learning.}
Our work is also related to clustering-based methods~\cite{asano2019self,bautista2016cliquecnn,caron2018deep,caron2019unsupervised,huang2019unsupervised,xie2016unsupervised,yang2016joint,zhuang2019local,gidaris2020learning,yan2020cluster}.
Caron~\etal~\cite{caron2018deep} show that $k$-means assignments can be used as pseudo-labels to learn visual representations.
This method scales to large uncurated dataset and can be used for pre-training of supervised networks~\cite{caron2019unsupervised}.
However, their formulation is not principled and recently, Asano~\etal~\cite{asano2019self} show how to cast the pseudo-label assignment problem as an instance of the optimal transport problem.
We consider a similar formulation to map representations to prototype vectors, but unlike~\cite{asano2019self} we keep the soft assignment produced by the Sinkhorn-Knopp algorithm~\cite{cuturi2013sinkhorn} instead of approximating it into a hard assignment.
Besides, unlike Caron~\etal~\cite{caron2018deep,caron2019unsupervised} and Asano~\etal~\cite{asano2019self}, we obtain online assignments which allows our method to scale gracefully to any dataset size.

\paragraph{Handcrafted pretext tasks.}
Many self-supervised methods manipulate the input data to extract a supervised signal in the form of a pretext task~\cite{doersch2015unsupervised,agrawal2015learning,jenni2018self,kim2018learning,larsson2016learning,mahendran2018cross,misra2016shuffle,pathak2017learning,pathak2016context,wang2015unsupervised,wang2017transitive,zhang2017split}.
We refer the reader to Jing~\etal~\cite{jing2019self} for an exhaustive and detailed review of this literature.
Of particular interest, Misra and van der Maaten~\cite{misra2019self} propose to encode the jigsaw puzzle task~\cite{noroozi2016unsupervised} as an invariant for contrastive learning.
Jigsaw tiles are non-overlapping crops with small resolution that cover only part (${\sim}20\%$) of the entire image area.
In contrast, our \texttt{multi-crop} strategy consists in simply sampling multiple random crops with two different sizes: a standard size and a smaller one.


\section{Method}

Our goal is to learn visual features in an online fashion without supervision.
To that effect, we propose an online clustering-based self-supervised method.
Typical clustering-based methods~\cite{asano2019self,caron2018deep} are offline in the sense that they alternate between a cluster assignment step where image features of the entire dataset are clustered, and a training step where the cluster assignments, \ie, ``codes'' are predicted for different image views.
Unfortunately, these methods are not suitable for online learning as they require multiple passes over the dataset to compute the image features necessary for clustering.
In this section, we describe an alternative where we enforce consistency between codes from different augmentations of the same image.
This solution is inspired by contrastive instance learning~\cite{wu2018unsupervised} as we do not consider the codes as a target, but only enforce consistent mapping between views of the same image.
Our method can be interpreted as a way of contrasting between multiple image views by comparing their cluster assignments instead of their features.

\begin{figure}[t]
\centering
\begin{tabular}{c | c}
\includegraphics[height=.2\linewidth]{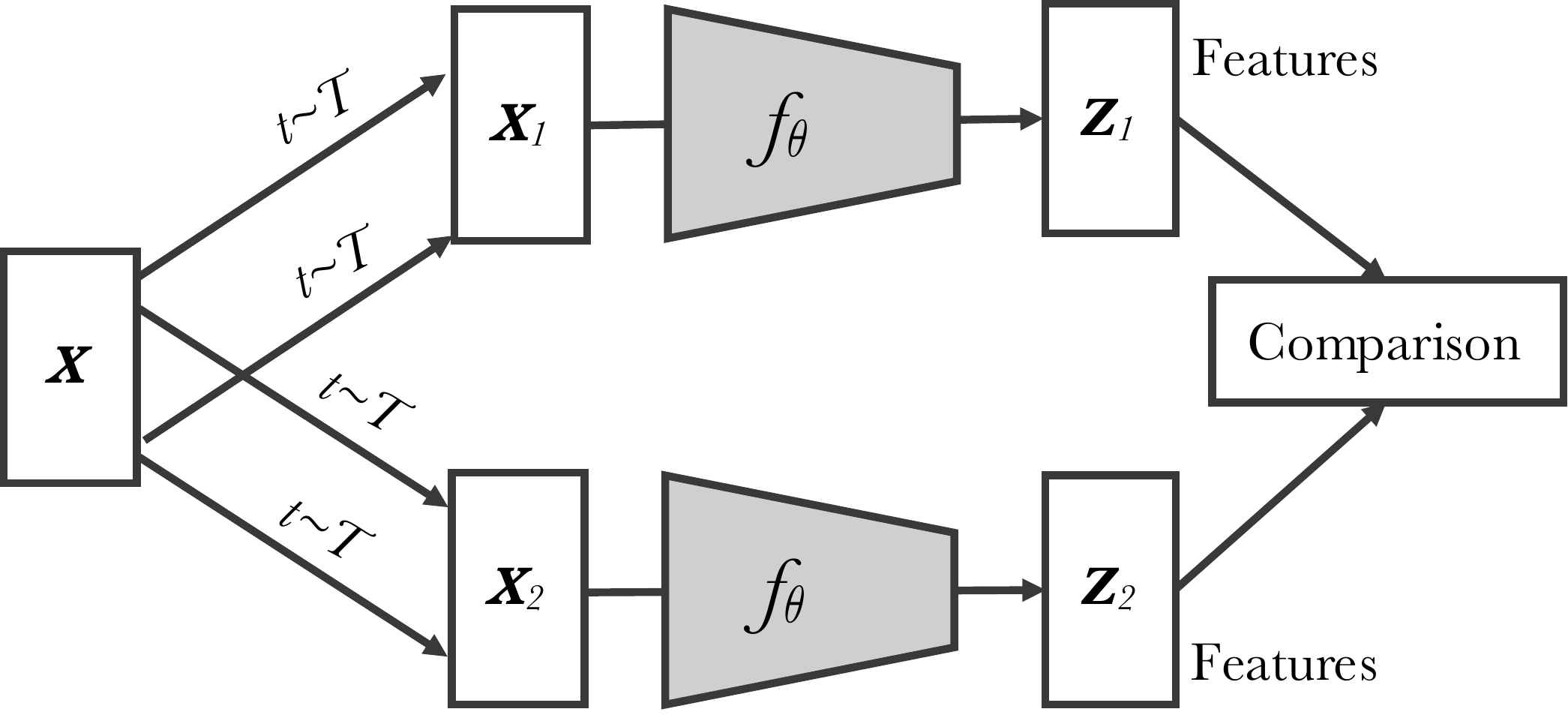} &
\includegraphics[height=.2\linewidth]{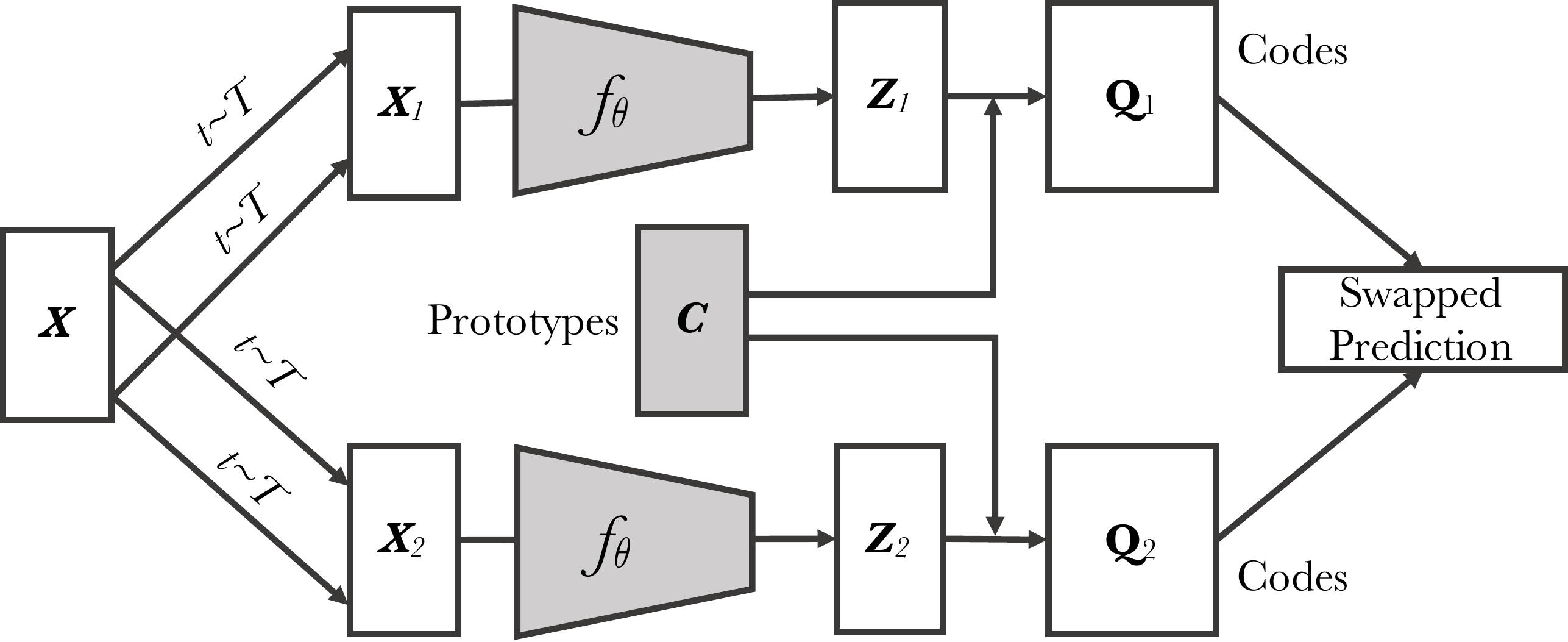} \\
\\
Contrastive instance learning & \OURSFULL (Ours)
\end{tabular}
   \caption{\textbf{Contrastive instance learning (left) \vs\ \OURS (right).}
In contrastive learning methods applied to instance classification, the features from different transformations of the same images are compared directly to each other.
In \OURS, we first obtain ``codes'' by assigning features to prototype vectors.
We then solve a ``swapped'' prediction problem wherein the codes obtained from one data augmented view are predicted using the other view.
Thus, \OURS does not directly compare image features.
Prototype vectors are learned along with the ConvNet parameters by backpropragation. 
}\label{fig:oto}
\end{figure}

More precisely, we compute a code from an augmented version of the image and predict this code from other augmented versions of the same image.
Given two image features $\mathbf{z}_t$ and $\mathbf{z}_s$ from two different augmentations of the same image, we compute their codes $\mathbf{q}_t$ and $\mathbf{q}_s$ by matching these features to a set of $K$ prototypes $\{\mathbf{c}_1,\dots, \mathbf{c}_K\}$.
We then setup a ``swapped'' prediction problem with the following loss function:
\begin{eqnarray}
L(\mathbf{z}_t, \mathbf{z}_s) & = &\ell(\mathbf{z}_t, \mathbf{q}_s) + \ell(\mathbf{z}_s, \mathbf{q}_t),
\label{eq:twoviews}
\end{eqnarray}
where the function~$\ell({\mathbf z},{\mathbf q})$ measures the fit between features ${\mathbf z}$ and a code ${\mathbf q}$, as detailed later.  
Intuitively, our method compares the features $\mathbf{z}_t$ and $\mathbf{z}_s$ using the intermediate codes $\mathbf{q}_t$ and $\mathbf{q}_s$. 
If these two features capture the same information, it should be possible to predict the code from the other feature.
A similar comparison appears in contrastive learning where features are compared directly~\cite{wu2018unsupervised}. 
In~\cref{fig:oto}, we illustrate the relation between contrastive learning and our method.


\subsection{Online clustering}
\label{sec:proto}

Each image $\mathbf{x}_n$ is transformed into an augmented view $\mathbf{x}_{nt}$ by applying a transformation $t$ sampled from the set $\mathcal{T}$ of image transformations.
The augmented view is mapped to a vector representation by applying a non-linear mapping $f_\theta$ to $\mathbf{x}_{nt}$.
The feature is then projected to the unit sphere,~\ie,~ $\mathbf{z}_{nt} = f_\theta(\mathbf{x}_{nt}) / \|f_\theta(\mathbf{x}_{nt})\|_2$.
We then compute a code $\mathbf{q}_{nt}$ from this feature by mapping $\mathbf{z}_{nt}$ to a set of $K$ trainable prototypes vectors,  $\{\mathbf{c}_1,\dots,\mathbf{c}_K\}$.
We denote by $\mathbf{C}$ the matrix whose columns are the $\mathbf{c}_1, \dots, \mathbf{c}_k$. 
We now describe how to compute these codes and update the prototypes online.

\paragraph{Swapped prediction problem.} 
The loss function in Eq.~(\ref{eq:twoviews}) has two terms that setup the ``swapped'' prediction problem of predicting the code $\mathbf{q}_t$ from the feature $\mathbf{z}_s$, and $\mathbf{q}_s$ from $\mathbf{z}_t$. 
Each term represents the cross entropy loss between the code and the probability obtained by taking a softmax of the dot products of $\mathbf{z}_i$ and all prototypes in $\mathbf{C}$, \ie, 
\begin{equation}
  \ell(\mathbf{z}_t, \mathbf{q}_s) = - \sum_{k} \mathbf{q}_s^{(k)} \log \mathbf{p}_t^{(k)}, \quad \text{where} \quad \mathbf{p}_t^{(k)} = \frac{ \exp \left ( \frac{1}{\tau} \mathbf{z}_t^\top \mathbf{c}_k \right ) }{\sum_{k'} \exp \left ( \frac{1}{\tau} \mathbf{z}_t^\top \mathbf{c}_{k'} \right ) }.
  \label{eq:loss}
\end{equation}
where $\tau$ is a temperature parameter~\cite{wu2018unsupervised}.
Taking this loss over all the images and pairs of data augmentations leads to the following loss function for the swapped prediction problem:
\begin{equation}
  - \frac{1}{N} \sum_{n=1}^N \sum_{s,t \sim \mathcal{T}} \left [ 
  \frac{1}{\tau} \mathbf{z}_{nt}^\top \mathbf{C} \mathbf{q}_{ns}
  + \frac{1}{\tau} \mathbf{z}_{ns}^\top \mathbf{C} \mathbf{q}_{nt}
  - \log \sum_{k=1}^K \exp \left ( \frac{\mathbf{z}_{nt}^\top \mathbf{c}_k}{\tau} \right )  
  - \log \sum_{k=1}^K \exp \left ( \frac{\mathbf{z}_{ns}^\top \mathbf{c}_k}{\tau} \right ) \right ] .
  \nonumber
\label{eq:swapploss2}
\end{equation}
This loss function is jointly minimized with respect to the prototypes $\mathbf{C}$ and the parameters $\theta$ of the image encoder $f_\theta$ used to produce the features $(\mathbf{z}_{nt})_{n,t}$.

\paragraph{Computing codes online.}

In order to make our method online, we compute the codes using only the image features within a batch.
Intuitively, as the prototypes $\mathbf{C}$ are used across different batches, \OURS clusters multiple instances to the prototypes.
We compute codes using the prototypes $\mathbf{C}$ such that all the examples in a batch are equally partitioned by the prototypes. 
This equipartition constraint ensures that the codes for different images in a batch are distinct, thus preventing the trivial solution where every image has the same code.
Given $B$ feature vectors $\mathbf{Z}=[\mathbf{z}_{1},\dots,\mathbf{z}_{B}]$, we are interested in mapping them to the prototypes $\mathbf{C}=[\mathbf{c}_1,\dots,\mathbf{c}_K]$.
We denote this mapping or codes by $\mathbf{Q} = [\mathbf{q}_1, \dots, \mathbf{q}_B]$, and optimize $\mathbf{Q}$ to maximize the similarity between the features and the prototypes , \ie, 
\begin{equation}
  \label{eq:assign}
  \max_{\mathbf{Q}\in\mathcal{Q}} \ \text{Tr} \left( \mathbf{Q}^\top \mathbf{C}^\top \mathbf{Z} \right) +  \varepsilon H(\mathbf{Q}),
\end{equation}
where $H$ is the entropy function, $H(\mathbf{Q}) = -\sum_{ij} \mathbf{Q}_{ij} \log \mathbf{Q}_{ij}$ and $\varepsilon$ is a parameter that controls the smoothness of the mapping.
We observe that a strong entropy regularization (\ie~using a high $\varepsilon$) generally leads to a trivial solution where all samples collapse into an unique representation and are all assigned uniformely to all prototypes.
Hence, in practice we keep $\varepsilon$ low.
Asano \etal~\cite{asano2019self}  enforce an equal partition by constraining the matrix $\mathbf{Q}$ to belong to the transportation polytope.
They work on the full dataset, and we propose to adapt their solution to work on minibatches by restricting the transportation polytope to the minibatch:
\begin{equation}
  \mathcal{Q} = \left \{\mathbf{Q}\in\mathbb{R}_+^{K\times B} ~|~\mathbf{Q} \mathbf{1}_B = \frac{1}{K} \mathbf{1}_K, \mathbf{Q}^\top \mathbf{1}_K = \frac{1}{B} \mathbf{1}_B \right \},
\end{equation}
where $\mathbf{1}_K$ denotes the vector of ones in dimension $K$.
These constraints enforce that on average each prototype is selected at least $\frac{B}{K}$ times in the batch.

Once a continuous solution $\mathbf{Q}^*$ to Prob.~(\ref{eq:assign}) is found, a discrete code can be obtained by using a rounding procedure~\cite{asano2019self}.
Empirically, we found that discrete codes work well when computing codes in an offline manner on the full dataset as in Asano~\etal~\cite{asano2019self}.
However, in the online setting where we use only minibatches, using the discrete codes performs worse than using the continuous codes.
An explanation is that the rounding needed to obtain discrete codes is a more aggressive optimization step than gradient updates.
While it makes the model converge rapidly, it leads to a worse solution.
We thus preserve the soft code $\mathbf{Q}^*$ instead of rounding it. 
These soft codes $\mathbf{Q}^*$ are the solution of Prob.~(\ref{eq:assign}) over the set $\mathcal{Q}$ and takes the form of a normalized exponential matrix~\cite{cuturi2013sinkhorn}:
\begin{equation}
\label{eq:qstar}
  \mathbf{Q}^*= \text{Diag}(\mathbf{u}) \exp\left(\frac{\mathbf{C}^\top \mathbf{Z}}{\varepsilon} \right) \text{Diag}(\mathbf{v}),
\end{equation}
where $\mathbf{u}$ and $\mathbf{v}$ are renormalization vectors in $\mathbb{R}^K$ and $\mathbb{R}^B$ respectively.
The renormalization vectors are computed using a small number of matrix multiplications using the iterative Sinkhorn-Knopp algorithm~\cite{cuturi2013sinkhorn}.
In practice, we observe that using only $3$ iterations is fast and sufficient to obtain good performance.
Indeed, this algorithm can be efficiently implemented on GPU, and the alignment of $4$K features to $3$K codes takes $35$ms in our experiments, see~\cref{sec:exp}.

\paragraph{Working with small batches.}
When the number $B$ of batch features is too small compared to the number of prototypes $K$, it is impossible to equally partition the batch into the $K$ prototypes.
Therefore, when working with small batches, we use features from the previous batches to augment the size of $\mathbf{Z}$ in Prob.~(\ref{eq:assign}).
Then, we only use the codes of the batch features in our training loss.
In practice, we store around $3$K features, \ie, in the same range as the number of code vectors.
This means that we only keep features from the last $15$ batches with a batch size of $256$, while contrastive methods typically need to store the last $65$K instances obtained from the last $250$ batches~\cite{he2019momentum}.

\begin{figure}[t!]
\begin{minipage}{0.45\linewidth}
\centering
    \begin{tabular}{@{} l l c c @{}}
      \toprule
      Method      & Arch. & Param. & Top1  \\
      \midrule
      Supervised  & \resnetfifty    & 24  & 76.5  \\
      \midrule
      Colorization~\cite{zhang2016colorful}     & \resnetfifty  & 24 & 39.6 \\
      Jigsaw~\cite{noroozi2016unsupervised}     & \resnetfifty  & 24 & 45.7 \\
      NPID~\cite{wu2018unsupervised}   & \resnetfifty  & 24 & 54.0   \\
      BigBiGAN~\cite{donahue2019large} & \resnetfifty  & 24 & 56.6   \\
      LA~\cite{zhuang2019local}        & \resnetfifty  & 24 & 58.8 \\
      NPID++~\cite{misra2019self}      & \resnetfifty  & 24 & 59.0   \\
      MoCo~\cite{he2019momentum}       & \resnetfifty  & 24 & 60.6   \\
      SeLa~\cite{asano2019self}        & \resnetfifty  & 24 & 61.5   \\
      PIRL~\cite{misra2019self}        & \resnetfifty  & 24 & 63.6   \\
      CPC v2~\cite{henaff2019data}     & \resnetfifty  & 24 & 63.8 \\
      PCL~\cite{li2020prototypical}     & \resnetfifty  & 24 & 65.9 \\
      SimCLR~\cite{chen2020simple}     & \resnetfifty  & 24 & 70.0  \\
      MoCov2~\cite{chen2020improved}   & \resnetfifty  & 24 & 71.1  \\
\midrule
      \OURS   		       & \resnetfifty  & 24 & \bf{75.3}        \\ 
      \bottomrule
 \end{tabular}
\end{minipage}
~~~~~\begin{minipage}{0.5\linewidth}
\centering
\includegraphics{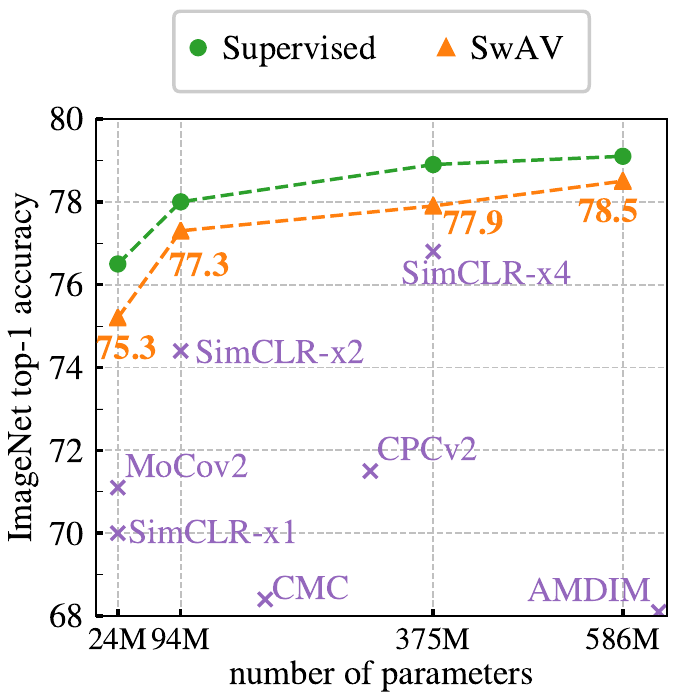} 
\end{minipage}
    \caption{
      \textbf{Linear classification on \ImNet.} 
Top-1 accuracy for linear models trained on frozen features from different self-supervised methods.
\textbf{(left)}
Performance with a standard ResNet-50.
\textbf{(right)} Performance as we multiply the width of a ResNet-50 by a factor $\times2$, $\times4$, and $\times5$.
}
    \label{fig:linear_inet}
\end{figure}

\subsection{Multi-crop: Augmenting views with smaller images}
\label{sec:multicrop}
As noted in prior works~\cite{chen2020simple,misra2019self}, comparing random crops of an image plays a central role by capturing information in terms of relations between parts of a scene or an object. 
Unfortunately, increasing the number of crops or ``views'' quadratically increases the memory and compute requirements. 
We propose a \texttt{multi-crop} strategy where we use two standard resolution crops and sample $V$ additional low resolution crops that cover only small parts of the image. 
Using low resolution images ensures only a small increase in the compute cost. 
Specifically, we generalize the loss of Eq~(\ref{eq:twoviews}):
\begin{equation}
L(\mathbf{z}_{t_1}, \mathbf{z}_{t_2}, \dots, \mathbf{z}_{t_{V + 2}}) = \sum_{i \in \{1, 2\}} \sum_{v=1}^{V + 2} \mathbf{1}_{v \neq i} \ell(\mathbf{z}_{t_v}, \mathbf{q}_{t_i}).
\end{equation}
Note that we compute codes using only the full resolution crops.
Indeed, computing codes for all crops increases the computational time and we observe in practice that it also alters the transfer performance of the resulting network.
An explanation is that using only partial information (small crops cover only small area of images) degrades the assignment quality.
Figure~\ref{fig:multicrop-exp} shows that \texttt{multi-crop} improves the performance of several self-supervised methods and is a promising augmentation strategy.

\section{Main Results}\label{sec:exp}
We analyze the features learned by \OURS by transfer learning on multiple datasets. 
We implement in \OURS the improvements used in SimCLR, \ie, LARS~\cite{you2017large}, cosine learning rate~\cite{loshchilov2016sgdr,misra2019self} and the MLP projection head~\cite{chen2020simple}. 
We provide the full details and hyperparameters for pretraining and transfer learning in the \appendixName.

\subsection{Evaluating the unsupervised features on ImageNet}
\label{sec:eval_on_imagenet}
We evaluate the features of a ResNet-50~\cite{he2016deep} trained with \OURS on ImageNet by two experiments: linear classification on frozen features and semi-supervised learning by finetuning with few labels.
When using frozen features (\cref{fig:linear_inet} left), \OURS outperforms the state of the art by $+4.2\%$ top-1 accuracy and is only $1.2\%$ below the performance of a fully supervised model.
Note that we train \OURS during $800$ epochs with large batches ($4096$).
We refer to~\cref{fig:multicrop-exp} for results with shorter trainings and to~\cref{fig:small_batches} for experiments with small batches.
On semi-supervised learning (\cref{tab:semi_sup}), \OURS outperforms other self-supervised methods and is on par with state-of-the-art semi-supervised models~\cite{sohn2020fixmatch}, despite the fact that \OURS is not specifically designed for semi-supervised learning.

\begin{table}[t]
  \centering
  \caption{
     \textbf{Semi-supervised learning on \ImNet with a ResNet-50.} 
We finetune the model with $1\%$ and $10\%$ labels and report top-1 and top-5 accuracies.
*:\footnotesize{\emph{ uses RandAugment~\cite{cubuk2019randaugment}.}}
    }
    \label{tab:semi_sup}
    \vspace{0.4em}
  \begin{tabular}{@{}ll@{}ll c c c c c @{}}
    \toprule
	  &&&& \multicolumn{2}{c}{1\% labels} && \multicolumn{2}{c}{10\% labels} \\
	  &&Method && Top-1 & Top-5 && Top-1 & Top-5 \\
    \midrule
	&&Supervised && 25.4 & 48.4 && 56.4 & 80.4 \\
    \midrule
	  \multirow{2}{*}{\shortstack[l]{\textit{Methods using} \\ \textit{label-propagation}}}
	   &&UDA~\cite{xie2020unsupervised} && - & - && \phantom{*}68.8* & \phantom{*}88.5* \\
	  &&FixMatch~\cite{sohn2020fixmatch} && - & - &&  \phantom{*}\textbf{71.5}* & \phantom{*}89.1* \\
    \midrule
	  \multirow{4}{*}{\shortstack[l]{\textit{Methods using} \\ \textit{self-supervision only}}}&&PIRL~\cite{misra2019self} && 30.7 & 57.2 && 60.4 & 83.8 \\
	  &&PCL~\cite{li2020prototypical} && - & 75.6 && - & 86.2 \\
	  &&SimCLR~\cite{chen2020simple} && 48.3 & 75.5 && 65.6 & 87.8 \\
\cmidrule{3-9}
	&&\OURS && \bf 53.9 & \bf 78.5 && 70.2 & \bf 89.9 \\
    \bottomrule
  \end{tabular}
\end{table}

\par \noindent \textbf{Variants of ResNet-50.} Figure~\ref{fig:linear_inet}~(right) shows the performance of multiple variants of ResNet-50 with different widths~\cite{kolesnikov2019revisiting}.
The performance of our model increases with the width of the model, and follows a similar trend to the one obtained with supervised learning.
When compared with concurrent work like SimCLR, we see that \OURS reduces the difference with supervised models even further.
Indeed, for large architectures, our method shrinks the gap with supervised training to $0.6\%$.

\begin{table}[h]
\centering
    \caption{
\textbf{Transfer learning on downstream tasks.}
Comparison between features from ResNet-50 trained on ImageNet with \OURS or supervised learning.
We consider two settings.
(1) Linear classification on top of frozen features. We report top-1 accuracy on all datasets except \VOCseven where we report mAP.
	(2) Object detection with finetuned features on \VOCseventwelve \texttt{trainval} using Faster R-CNN~\cite{ren2015faster} and on COCO~\cite{lin2014microsoft} using Mask R-CNN~\cite{he2017mask} or DETR~\cite{carion2020end}.
We report the most standard detection metrics for these datasets: $\text{AP}_{50}$ on \VOCseventwelve and $\text{AP}$ on COCO.
    }
    \label{fig:frozen_and_detect}
    \vspace{.4em}
  \setlength{\tabcolsep}{3.7pt}
    \begin{tabular}{ @{} l ccc c ccc @{} }
      \toprule
       & \multicolumn{3}{c}{Linear Classification} &~~~~& \multicolumn{3}{c}{Object Detection}\\
\cmidrule{2-4}\cmidrule{6-8}
	    &\Places & \VOCseven & \iNat && \VOCseventwelve & COCO & COCO \\
	    & &  &  && \scriptsize(Faster R-CNN R50-C4) & \scriptsize(Mask R-CNN R50-FPN) & \scriptsize(DETR) \\
      \midrule
	    Supervised & 53.2  & 87.5 & 46.7 && 81.3 & 39.7 & 40.8 \\
      \midrule
	    \OURS  & \bf 56.7 & \bf 88.9 & \bf 48.6 && \bf 82.6 & \bf 41.6 & \bf 42.1 \\
      \bottomrule
    \end{tabular}
\end{table}

\subsection{Transferring unsupervised features to downstream tasks}
\label{sec:downstream}
We test the generalization of ResNet-50 features trained with \OURS on \ImNet (without labels) by transferring to several downstream vision tasks.
In~\cref{fig:frozen_and_detect}, we compare the performance of~\OURS features with ImageNet supervised pretraining.
First, we report the linear classification performance on the \Places~\cite{zhou2014learning}, \VOCseven~\cite{everingham2010pascal}, and iNaturalist2018~\cite{van2018inaturalist} datasets.
Our method outperforms supervised features on all three datasets.
Note that \OURS is the first self-supervised method to surpass ImageNet supervised features on these datasets.
Second, we report network finetuning on object detection on \VOCseventwelve using Faster R-CNN~\cite{ren2015faster} with a R50-C4 backbone and on COCO~\cite{lin2014microsoft} using Mask R-CNN~\cite{he2017mask} with a R50-FPN backbone and finally using DETR detector~\cite{carion2020end}.
DETR is a recent object detection framework that reaches competitive performance with Faster R-CNN while being conceptually simpler and trainable end-to-end.
Interestingly, unlike Faster R-CNN~\cite{ren2015faster}, using a pretrained backbone for DETR is crucial to obtain good results compared to training from scratch~\cite{carion2020end}.
In~\cref{fig:frozen_and_detect}, we show that \OURS outperforms the supervised pretrained model on both \VOCseventwelve and COCO datasets.
Note that this is line with previous works that also show that self-supervision can outperform supervised pretraining on object detection~\cite{misra2019self,he2019momentum,gidaris2020learning}.
We report more detection evaluation metrics and results from other self-supervised methods in the \appendixName.
Overall, our \OURS ResNet-50 model surpasses supervised ImageNet pretraining on all the considered transfer tasks and datasets.
We have released this model so other researchers might also benefit by replacing the ImageNet supervised network with our model.

\subsection{Training with small batches}
We train \OURS with small batches of $256$ images on $4$ GPUs and compare with MoCov2 and SimCLR trained in the same setup.
In \cref{fig:small_batches}, we see that \OURS maintains state-of-the-art performance even when trained in the small batch setting.
Note that \OURS only stores a queue of $3,840$ features.
In comparison, to obtain good performance, MoCov2 needs to store $65,536$ features while keeping an additional momentum encoder network.
When \OURS is trained using $2\!\times\!160+4\!\times\!96$ crops, \OURS has a running time $1.2\times$ higher than SimCLR with $2\!\times\!224$ crops and is around $1.4\times$ slower than MoCov2 due to the additional back-propagation~\cite{chen2020improved}.
Hence, one epoch of MoCov2 or SimCLR is faster in wall clock time than one of \OURS, but these methods need more epochs for good downstream performance.
Indeed, as shown in~\cref{fig:small_batches}, \OURS learns much faster and reaches higher performance in $4\times$ fewer epochs: $72\%$ after $200$ epochs ($102$ hours) while MoCov2 needs $800$ epochs to achieve $71.1\%$.
Increasing the resolution and the number of epochs, \OURS reaches $74.3\%$ with a small batch size, a small number of stored features and no momentum encoder.
Finally, note that \OURS could be combined with a momentum mechanism and a large queue~\cite{he2019momentum}; we leave these explorations to future work.\\

\begin{table}[t]
\centering
\caption{
\textbf{Training in small batch setting.} 
Top-1 accuracy on ImageNet with a linear classifier trained on top of frozen features from a ResNet-50.
All methods are trained with a batch size of $256$.
We also report the number of stored features, the type of cropping used and the number of epochs.
}\label{fig:small_batches} 
\vspace{.3em}
    \begin{tabular}{@{} l  c c c c c  c @{}}
      \toprule
    Method      & Mom. Encoder & Stored Features & \texttt{multi-crop} & epoch & batch  &  Top-1  \\
      \midrule
	    SimCLR 	  &   & $0$ & $2\!\times\!224$ & $200$ & $256$   & $61.9$ \\
	    MoCov2      & \checkmark & $65,536$ & $2\!\times\!224$ & $200$ & $256$   & $67.5$  \\
      MoCov2      & \checkmark & $65,536$  & $2\!\times\!224$ & $800$ & $256$   & $71.1$  \\
      \midrule
      \OURS 	  &  &  $3,840$  & $2\!\times\!160$ + $4\!\times\!96$ & $200$ & $256$ & $72.0$ \\
      \OURS 	  &  &  $3,840$  & $2\!\times\!224$ + $6\!\times\!96$ & $200$ & $256$ & $72.7$ \\
      \OURS 	  &  &  $3,840$  & $2\!\times\!224$ + $6\!\times\!96$ & $400$ & $256$ & $\mathbf{74.3}$ \\
      \bottomrule
    \end{tabular}
\end{table}

\section{Ablation Study}
\label{sec:ablation}

\subsection{Clustering-based self-supervised learning}

\paragraph{Improving prior clustering-based approaches.}
In this section, we re-implement and improve previously published clustering-based models in order to assess if they can compete with recent contrastive methods such as SimCLR.
In particular, we consider two clustering-based models: DeepCluster-v2 and SeLa-v2, which are obtained by applying various training improvements introduced in other self-supervised learning papers to DeepCluster~\cite{caron2018deep} and SeLa~\cite{asano2019self}.
Among these improvements are the use of stronger data augmentation~\cite{chen2020simple}, MLP projection head~\cite{chen2020simple}, cosine learning rate schedule~\cite{misra2019self}, temperature parameter~\cite{wu2018unsupervised}, memory bank~\cite{wu2018unsupervised}, multi-clustering~\cite{asano2019self}, etc.
Full implementation details can be found in the \appendixName.
Besides, we also improve DeepCluster model by introducing explicit comparisons to k-means centroids, which increase stability and performance.
Indeed, a main issue in DeepCluster is that there is no correspondance between two consecutive cluster assignments.
Hence, the final classification layer learned for an assignment becomes irrelevant for the following one and thus needs to be re-initialized from scratch at each epoch.
This considerably disrupts the convnet training.
In DeepCluster-v2, instead of learning a classification layer predicting the cluster assignments, we perform explicit comparison between features and centroids.

\begin{figure}[t]
  \begin{minipage}{0.5\linewidth}
    \begin{tabular}{ @{} l c c c  @{}}
      \toprule
                    & \multicolumn{2}{c}{Top-1} & $\Delta$ \\
                    \cmidrule{2-3}
      Method        & 2x224 & 2x160+4x96 &  \\
      \midrule
      Supervised    & $76.5$ & $76.0$ & $-0.5$\\
      \midrule
  \multicolumn{4}{l}{\textit{Contrastive-instance approaches}}\\
      SimCLR        & $68.2$ & $70.6$ & $+2.4$\\
      \midrule
  \multicolumn{4}{l}{\textit{Clustering-based approaches}}\\
      SeLa-v2        & $67.2$ & $71.8$ & $+4.6$\\
      DeepCluster-v2 & $70.2$ & $74.3$ & $+4.1$\\
      \OURS         & $70.1$ & $74.1$ & $+4.0$\\
      \bottomrule
    \end{tabular}
  \end{minipage}
  \begin{minipage}{0.5\linewidth}
    \centering
    \includegraphics{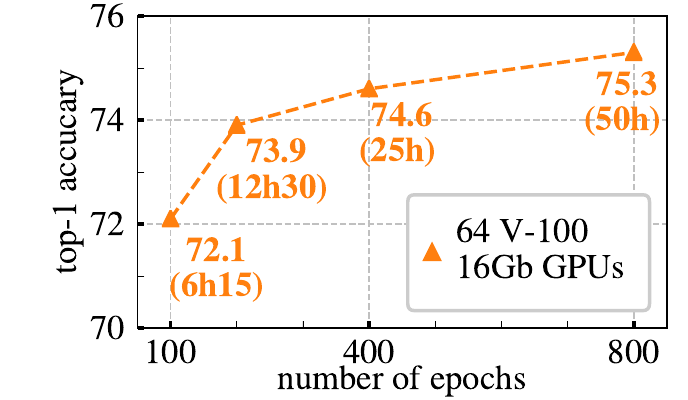}
  \end{minipage}
  \vspace{-0.3em}
  \caption{
    Top-1 accuracy on ImageNet with a linear classifier trained on top of frozen features from a ResNet-50.
    \textbf{(left) Comparison between clustering-based and contrastive instance methods and impact of multi-crop.} 
    Self-supervised methods are trained for $400$ epochs and supervised models for $200$ epochs.
    \textbf{(right) Performance as a function of epochs.}
    We compare \OURS models trained with different number of epochs and report their running time based on our implementation.
  }
  \label{fig:multicrop-exp}
\end{figure}

\paragraph{Comparing clustering with contrastive instance learning.}
In~\cref{fig:multicrop-exp}~(left), we make a best effort fair comparison between clustering-based and contrastive instance (SimCLR) methods by implementating these methods with the same data augmentation, number of epochs, batch-sizes, etc.
In this setting, we observe that \OURS and DeepCluster-v2 outperform SimCLR by $2\%$ without \texttt{multi-crop} and by $3.5\%$ with \texttt{multi-crop}.
This suggests the learning potential of clustering-based methods over instance classification.

\paragraph{Advantage of \OURS compared to DeepCluster-v2.}
In~\cref{fig:multicrop-exp}~(left), we observe that \OURS performs on par with DeepCluster-v2.
In addition, we train DeepCluster-v2 in SwAV best setting (800 epochs - 8 crops) and obtain $75.2\%$ top-1 accuracy on ImageNet (versus $75.3\%$ for \OURS).
However, unlike \OURS, DeepCluster-v2 is not online which makes it impractical for extremely large datasets (\cref{sec:uncurated}).
For billion scale trainings for example, a single pass on the dataset is usually performed~\cite{he2019momentum}.
DeepCluster-v2 cannot be trained for only one epoch since it works by performing several passes on the dataset to regularly update centroids and cluster assignments for each image.

As a matter of fact, DeepCluster-v2 can be interpreted as a special case of our proposed swapping mechanism: swapping is done across epochs rather than within a batch.
Given a crop of an image DeepCluster-v2 predicts the assignment of another crop, which was obtained at the previous epoch.
\OURS swaps assignments directly at the batch level and can thus work online.

\subsection{Applying the multi-crop strategy to different methods}
In~\cref{fig:multicrop-exp}~(left), we report the impact of applying our \texttt{multi-crop} strategy on the performance of a selection of other methods.
Details of how we apply \texttt{multi-crop} to SimCLR loss can be found in the \appendixName.
We see that the \texttt{multi-crop} strategy consistently improves the performance for all the considered methods by a significant margin of $2\!-\!4\%$ top-1 accuracy.
Interestingly, \texttt{multi-crop} seems to benefit more clustering-based methods than contrastive methods.
We note that \texttt{multi-crop} does not improve the supervised model.

\subsection{Impact of longer training}
In~\cref{fig:multicrop-exp} (right), we show the impact of the number of training epochs on performance for \OURS with \texttt{multi-crop}.
We train separate models for $100$, $200$, $400$ and $800$ epochs and report the top-1 accuracy on ImageNet using the linear classification evaluation.
We train each ResNet-50 on $64$ V100 16GB GPUs and a batch size of $4096$.
While \OURS benefits from longer training, it already achieves strong performance after $100$ epochs, \ie, $72.1\%$ in 6h15.

\subsection{Unsupervised pretraining on a large uncurated dataset}
\label{sec:uncurated}
\begin{figure}[t]
\begin{minipage}{0.5\linewidth}
\centering

    \begin{tabular}{@{} l  c c @{}}
\toprule
Method  & Frozen & Finetuned \\
\midrule
Random  & 15.0 & 76.5\\
\midrule
MoCo  & - & \phantom{*}77.3*\\
SimCLR  & 60.4 & 77.2 \\
\OURS  & \bf 66.5 & \bf 77.8 \\
      \bottomrule
    \end{tabular}
\end{minipage}
\begin{minipage}{0.5\linewidth}
\centering
\includegraphics{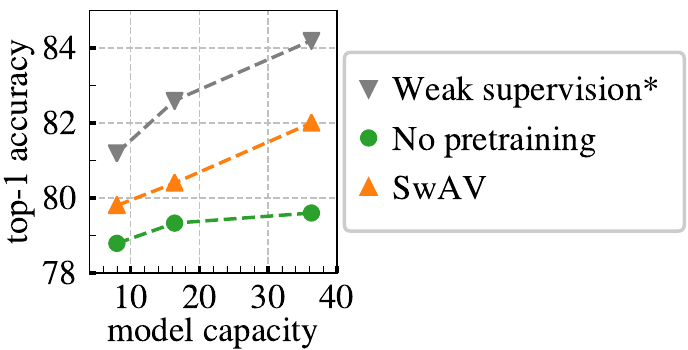} 
\end{minipage}
\vspace{-0.08in}
\caption{\textbf{Pretraining on uncurated data.}
Top-1 accuracy on ImageNet for pretrained models on an uncurated set of 1B random Instagram images.
\textbf{(left)} We compare ResNet-50 pretrained with either SimCLR or \OURS on two downstream tasks:
linear classification on frozen features or finetuned features.
\textbf{(right)}
Performance of finetuned models as we increase the capacity of a ResNext following~\cite{mahajan2018exploring}. 
The capacity is provided in billions of Mult-Add operations.\\
*: \footnotesize{\emph{pretrained on a curated set of $1$B Instagram images filtered with $1.5$k hashtags similar to ImageNet classes.}}
}
    \label{fig:insta}
\end{figure}

We evaluate SwAV on random, uncurated images that have different properties from ImageNet which allows us to test if our online clustering scheme and multi-crop augmentation work out of the box.
In particular, we pretrain \OURS on an uncurated dataset of 1 billion random public non-EU images from Instagram.
We test if \OURS can serve as a pretraining method for supervised learning.
In~\cref{fig:insta}~(left), we measure the performance of ResNet-50 models when transferring to ImageNet with frozen or finetuned features.
We report the results from He~\etal~\cite{he2019momentum} but note that their setting is different.
They use a curated set of Instagram images, filtered by hashtags similar to ImageNet labels~\cite{mahajan2018exploring}.
We compare \OURS with a randomly initialized network and with a network pretrained on the same data using SimCLR.
We observe that \OURS maintains a similar gain of $6\%$ over SimCLR as when pretrained on ImageNet (\cref{fig:linear_inet}), showing that our improvements do not depend on the data distribution.
We also see that pretraining with \OURS on random images significantly improves over training from scratch on ImageNet ($+1.3\%$)~\cite{caron2019unsupervised,he2019momentum}.
This result is in line with Caron~\etal~\cite{caron2019unsupervised} and He~\etal~\cite{he2019momentum}.
In~\cref{fig:insta}~(right), we explore the limits of pretraining as we increase the model capacity.
We consider the variants of the ResNeXt architecture~\cite{xie2017aggregated} as in Mahajan \etal~\cite{mahajan2018exploring}.
We compare \OURS with supervised models trained from scratch on ImageNet.
For all models, \OURS outperforms training from scratch by a significant margin showing that it can take advantage of the increased model capacity.
For reference, we also include the results from Mahajan \etal~\cite{mahajan2018exploring} obtained with a weakly-supervised model pretrained by predicting hashtags filtered to be similar to ImageNet classes.
Interestingly, \OURS performance is strong when compared to this topline despite not using any form of supervision or filtering of the data. 

\section{Discussion}
Self-supervised learning is rapidly progressing compared to supervised learning, even surpassing it on transfer learning, even though the current experimental settings are designed for supervised learning.
In particular, architectures have been designed for supervised tasks, and it is not clear if the same models would emerge from exploring architectures with no supervision.
Several recent works have shown that exploring architectures with search~\cite{liu2020labels} or pruning~\cite{caron2020pruning} is possible without supervision, and we plan to evaluate the ability of our method to guide model explorations.

\paragraph{Acknowledgement.}
We thank Nicolas Carion, Kaiming He, Herve Jegou, Benjamin Lefaudeux, Thomas Lucas, Francisco Massa, Sergey Zagoruyko, and the rest of Thoth and FAIR teams for their help and fruitful discussions.
Julien Mairal was funded by the ERC grant number 714381 (SOLARIS project) and by ANR 3IA MIAI@Grenoble Alpes (ANR-19-P3IA-0003).

\bibliographystyle{splncs04}
\bibliography{egbib}

\newpage
\renewcommand{\thesubsection}{\Alph{subsection}}
\newcommand\tab[1][5mm]{\hspace*{#1}}

\subsection{Implementation Details}
\label{ap:implementation}
In this section, we provide the details and hyperparameters for \OURS pretraining and transfer learning.
Our code is publicly available at \url{https://github.com/facebookresearch/swav}.

\subsubsection{Implementation details of \OURS training}
\label{ap:implementation_swav}
First, we provide a pseudo-code for \OURS training loop using two crops in Pytorch style:
\\
\\
\com{\# C: prototypes (DxK)}
\\
\com{\# model: convnet + projection head}
\\
\com{\# temp: temperature}
\\
\\
\code{for x in loader:}  \com{\# load a batch x with B samples}
\\
\tab \code{x\_t = t(x)}  \com{\# t is a random augmentation}
\\
\tab \code{x\_s = s(x)}  \com{\# s is a another random augmentation}
\\
\\
\tab \code{z = model(cat(x\_t, x\_s))}  \com{\# embeddings: 2BxD}
\\
\\
\tab \code{scores = mm(z, C)}  \com{\# prototype scores: 2BxK}
\\
\tab \code{scores\_t = scores[:B]}
\\
\tab \code{scores\_s = scores[B:]}
\\
\\
\tab \com{\# compute assignments}
\\
\tab \code{with torch.no\_grad():}
\\
\tab \tab \code{q\_t = sinkhorn(scores\_t)}
\\
\tab \tab \code{q\_s = sinkhorn(scores\_s)}
\\
\\
\tab \com{\# convert scores to probabilities}
\\
\tab \code{p\_t = Softmax(scores\_t / temp)}
\\
\tab \code{p\_s = Softmax(scores\_s / temp)}
\\
\\
\tab \com{\# swap prediction problem}
\\
\tab \code{loss = - 0.5 * mean(q\_t * log(p\_s) + q\_s * log(p\_t))}
\\
\\
\tab \com{\# SGD update: network and prototypes}
\\
\tab \code{loss.backward()}
\\
\tab \code{update(model.params)}
\\
\tab \code{update(C)}
\\
\\
\tab \com{\# normalize prototypes}
\\
\tab \code{with torch.no\_grad():}
\\
\tab \tab \code{C = normalize(C, dim=0, p=2)}
\\
\\
\com{\# Sinkhorn-Knopp}
\\
\code{def sinkhorn(scores, eps=0.05, niters=3):}
\\
\tab \code{Q = exp(scores / eps).T}
\\
\tab \code{Q /= sum(Q)}
\\
\tab \code{K, B = Q.shape}
\\
\tab \code{u, r, c = zeros(K), ones(K) / K, ones(B) / B}
\\
\tab \code{for \_ in range(niters):}
\\
\tab \tab \code{u = sum(Q, dim=1)}
\\
\tab \tab \code{Q *= (r / u).unsqueeze(1)}
\\
\tab \tab \code{Q *= (c / sum(Q, dim=0)).unsqueeze(0)}
\\
\tab \code{return (Q / sum(Q, dim=0, keepdim=True)).T}
\\
\\
Most of our training hyperparameters are directly taken from SimCLR work~\cite{chen2020simple}.
We train \OURS with stochastic gradient descent using large batches of $4096$ different instances.
We distribute the batches over $64$ V$100$ $16$Gb GPUs, resulting in each GPU treating $64$ instances.
The temperature parameter $\tau$ is set to $0.1$ and the Sinkhorn regularization parameter $\varepsilon$ is set to $0.05$ for all runs.
We use a weight decay of $10^{-6}$, LARS optimizer~\cite{you2017large} and a learning rate of $4.8$ which is linearly ramped up during the first $10$ epochs.
After warmup, we use the cosine learning rate decay~\cite{loshchilov2016sgdr,misra2019self} with a final value of $0.0048$.
To help the very beginning of the optimization, we freeze the prototypes during the first epoch of training.
We synchronize batch-normalization layers across GPUs using the optimized implementation with kernels through CUDA/C-v2 extension from \texttt{apex}\footnote{\scriptsize\url{github.com/NVIDIA/apex}}.
We also use \texttt{apex} library for training with mixed precision~\cite{micikevicius2017mixed}.
Overall, thanks to these training optimizations (mixed precision, kernel batch-normalization and use of large batches~\cite{goyal2017accurate}), $100$ epochs of training for our best \OURS model take approximately $6$ hours (see~\cref{tab:cost}).
Similarly to previous works~\cite{chen2020simple,chen2020improved,li2020prototypical}, we use a projection head on top of the convnet features that consists in a $2$-layer multi-layer perceptron (MLP) that projects the convnet output to a $128$-D space.

Note that \OURS is more suitable for a multi-node distributed implementation compared to contrastive approaches SimCLR or MoCo.
The latter methods require sharing the feature matrix across all GPUs at every batch which might become a bottleneck when distributing across many GPUs.
On the contrary, \OURS requires sharing only matrix normalization statistics (sum of rows and columns) during the Sinkhorn algorithm.

\subsubsection{Data augmentation used in \OURS}
\begin{figure}[h]
\includegraphics[width=0.98\linewidth]{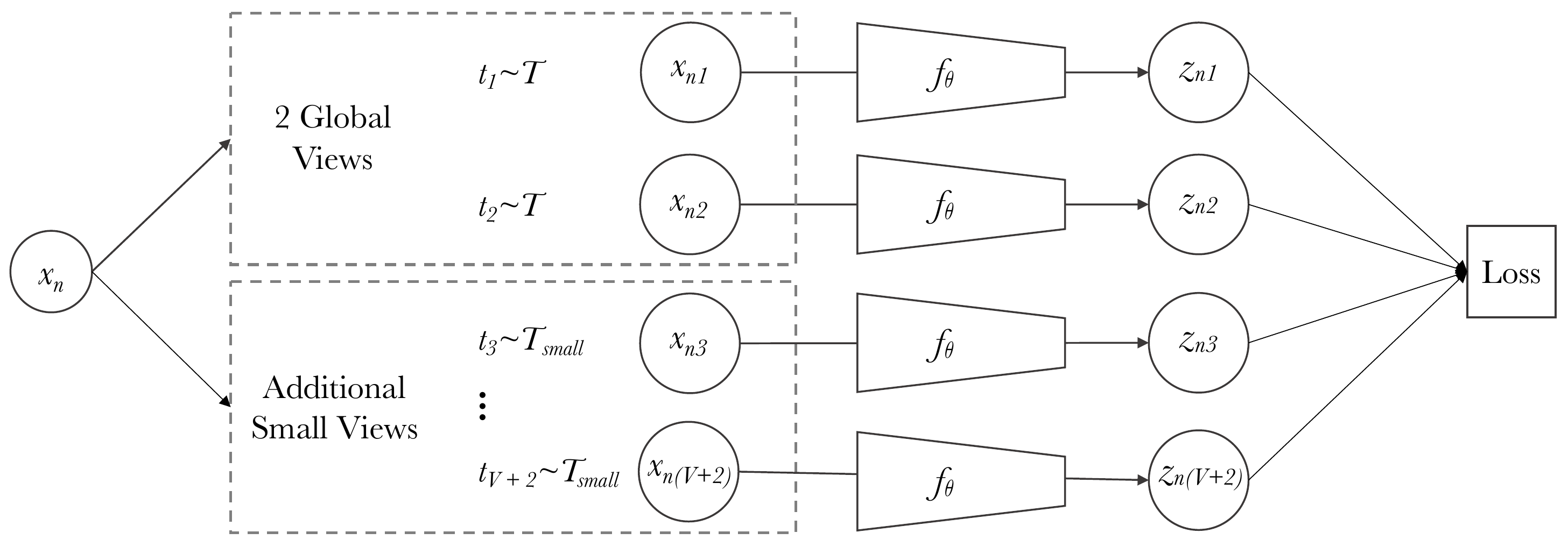}
	\caption{\textbf{Multi-crop}: the image $x_n$ is transformed into $V + 2$ views: two global views and $V$ small resolution zoomed views.}
\label{fig:multicrop}
\end{figure}

We obtain two different views from an image by performing crops of random sizes and aspect ratios.
Specifically we use the \texttt{RandomResizedCrop} method from \texttt{torchvision.transforms} module of PyTorch with the following scaling parameters: \texttt{s=(0.14, 1)}.
Note that we sample crops in a narrower range of scale compared to the default \texttt{RandomResizedCrop} parameters.
Then, we resize both full resolution views to $224 \times 224$ pixels, unless specified otherwise (we use $160 \times 160$ resolutions in some of our experiments).
Besides, we obtain $V$ additional views by cropping small parts in the image.
To do so, we use the following \texttt{RandomResizedCrop} parameters: \texttt{s=(0.05, 0.14)}.
We resize the resulting crops to $96 \times 96$ resolution.
Note that we always deal with resolutions that are divisible by $32$ to avoid roundings in the ResNet-$50$ pooling layers.
Finally, we apply random horizontal flips, color distortion and Gaussian blur to each resulting crop, exactly following the SimCLR implementation~\cite{chen2020simple}.
An illustration of our \texttt{multi-crop} augmentation strategy can be viewed in~\cref{fig:multicrop}.

\subsubsection{Implementation details of linear classification on \ImNet with ResNet-50}
We obtain $75.3$ top-1 accuracy on ImageNet by training a linear classifier on top of frozen final representations ($2048$-D) of a ResNet-$50$ trained with \OURS.
This linear layer is trained during $100$ epochs, with a learning rate of $0.3$ and a weight decay of $10^{-6}$.
We use cosine learning rate decay and a batch size of $256$.
We use standard data augmentations, i.e., cropping of random sizes and aspect ratios (default parameters of \texttt{RandomResizedCrop}) and random horizontal flips.

\subsubsection{Implementation details of semi-supervised learning (finetuning with 1\% or 10\% labels)}
We finetune with either 1\% or 10\% of ImageNet labeled images a ResNet-$50$ pretrained with \OURS.
We use the 1\% and 10\% splits specified in the official code release of SimCLR.
We mostly follow hyperparameters from PCL~\cite{li2020prototypical}: we train during $20$ epochs with a batch size of $256$, we use distinct learning rates for the convnet weights and the final linear layer, and we decay the learning rates by a factor $0.2$ at epochs $12$ and $16$.
We do not apply any weight decay during finetuning.
For 1\% finetuning, we use a learning rate of $0.02$ for the trunk and $5$ for the final layer.
For 10\% finetuning, we use a learning rate of $0.01$ for the trunk and $0.2$ for the final layer.

\subsubsection{Implementation details of transfer learning on downstream tasks}
\par \noindent \textbf{Linear classifiers.} We mostly follow PIRL~\cite{misra2019self} for training linear models on top of representations given by a ResNet-50 pretrained with \OURS.
On \VOCseven, all images are resized to 256 pixels along the shorter side, before taking a $224 \times 224$ center crop.
Then, we train a linear SVM with LIBLINEAR~\cite{fan2008liblinear} on top of corresponding global average pooled final representations ($2048$-D).
For linear evaluation on other datasets (\Places and \iNat), we train linear models with stochastic gradient descent using a batch size of $256$, a learning rate of $0.01$ reduced by a factor of $10$ three times (equally spaced intervals), weight decay of $0.0001$ and momentum of $0.9$.
On \Places, we train the linear models for $28$ epochs and on \iNat for $84$ epochs. We report the top-1 accuracy computed using the $224 \times 224$ center crop on the validation set.

\par \noindent \textbf{Object Detection on \VOCseventwelve.} We use a Faster R-CNN~\cite{ren2015faster} model as implemented in Detectron2~\cite{wu2019detectron2} and follow the finetuning protocol from He~\etal~\cite{he2019momentum} making the following changes to the hyperparameters -- our initial learning rate is $0.1$ which is warmed with a slope (\texttt{WARMUP\_FACTOR} flag in Detectron2) of $0.333$ for $1000$ iterations.
Other training hyperparamters are kept exactly the same as in He~\etal~\cite{he2019momentum}, \ie, batchsize of $16$ across $8$ GPUs, training for $24$K iterations on \VOCseventwelve \texttt{trainval} with the learning rate reduced by a factor of $10$ after $18$K and $22$K iterations, using SyncBatchNorm to finetune BatchNorm parameters, and adding an extra BatchNorm layer after the \texttt{res5} layer (\texttt{Res5ROIHeadsExtraNorm} head in Detectron2).
We report results on \VOCseven test set averaged over $5$ independant runs.

\par \noindent \textbf{Object Detection on COCO.}
We test the generalization of our ResNet-50 features trained on \ImNet with \OURS by transferring them to object detection on COCO dataset~\cite{lin2014microsoft} with DETR framework~\cite{carion2020end}.
DETR is a recent object detection framework that relies on a transformer encoder-decoder architecture.
It reaches competitive performance with Faster R-CNN while being conceptually simpler and trainable end-to-end.
Interestingly, unlike other frameworks~\cite{he2019rethinking}, current results with DETR have shown that using a pretrained backbone is crucial to obtain good results compared to training from scratch.
Therefore, we investigate if we can boost DETR performance by using features pretrained on ImageNet with \OURS instead of standard supervised features.
We also evaluate features from MoCov2~\cite{chen2020improved} pretraining.
We train DETR during $300$ epochs with AdamW, we use a learning rate of $10^{-4}$ for the transformer and apply a weight decay of $10^{-4}$.
We select for each method the best learning rate for the backbone among the following three values: $10^{-5}$, $5 \times 10^{-5}$ and $10^{-4}$.
We decay the learning rates by a factor $0.1$ after $200$ epochs.

\subsubsection{Implementation details of training with small batches of 256 images}
We start using a queue composed of the feature representations from previous batches after $15$ epochs of training.
Indeed, we find that using the queue before $15$ epochs disturbs the convergence of the model since the network is changing a lot from an iteration to another during the first epochs.
We simulate large batches of size $4096$ by storing the last $15$ batches, that is $3,840$ vectors of dimension $128$.
We use a weight decay of $10^{-6}$, LARS optimizer~\cite{you2017large} and a learning rate of $0.6$.
We use the cosine learning rate decay~\cite{loshchilov2016sgdr} with a final value of $0.0006$.

\subsubsection{Implementation details of ablation studies}
In our ablation studies (results in Table $5$ of the main paper for example), we choose to follow closely the data augmentation used in concurrent work SimCLR.
This allows a fair comparison and importantly, isolates the effect of our contributions.
In practice, it means that we use the default parameters of the random crop method (\texttt{RandomResizedCrop}), \texttt{s=(0.08, 1)} instead of \texttt{s=(0.14, 1)}, when sampling the two large resolution views.

\subsubsection{SimCLR loss with multi-crop augmentation}
When implementing SimCLR with \texttt{multi-crop} augmentation, we have to deal with several positive pairs formed by an anchor features and the different crops coming from the same instance.
We denote by $B$ the total number of unique dataset instances in the batch and by $M$ the number of crops per instance.
For example, in the case of 2x160+4x96 crops, we have $M=6$ crops per instance.
We call $N = B \times M$ the effective total number of crops in the batch.
Overall, we minimize the following loss
\begin{equation}
\mathcal{L} = - \frac{1}{N} \frac{1}{M - 1} \sum_{i=1}^N \sum_{v^+ \in \{v_i^+\}} \log{\frac{\exp{z_i^T v^+ / \tau}}{\exp{z_i^T v^+ / \tau} + \sum_{v^- \in \{v_i^-\}} \exp{z_i^T v^- / \tau}}}.
\end{equation}
For a feature representation $z_i$, the corresponding set of positive examples $\{v_i^+\}$ is formed by the representations of the other crops from the same instance.
The set of negatives $\{v_i^-\}$ is formed by the representations of all crops in the same batch except ones coming from the same instance as $x_i$.
Note that this extension of SimCLR loss with several pairs of positive is similar to the one used in Khosla~\etal~\cite{khosla2020supervised}.

\subsection{Additional Results}
\label{ap:results}

\subsubsection{Running times}
\begin{figure}[h]
\begin{minipage}{0.5\linewidth}
\centering
\includegraphics[width=\linewidth]{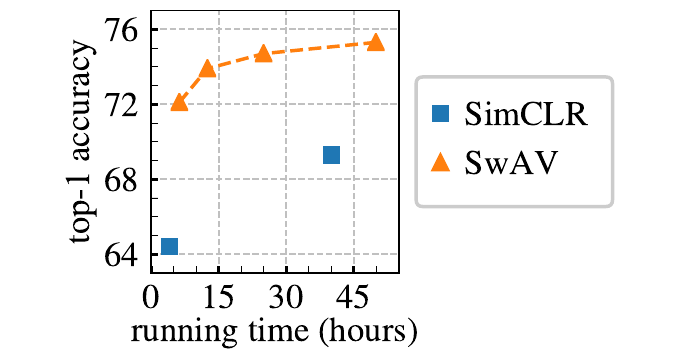}
\end{minipage}
\begin{minipage}{0.5\linewidth}
\caption{
\textbf{Influence of longer training.}
Top-1 ImageNet accuracy for linear models trained on frozen features.
We report \OURS performance for different training lengths measured in hours based on our implementation.
We train each ResNet-$50$ models on $64$ V100 16GB GPUs with a batch size of $4096$ (see \cref{ap:implementation_swav} for implementation details).
} 
\label{fig:cost}
\end{minipage}
\end{figure}

In~\cref{tab:cost}, we report compute and GPU memory requirements based on our implementation for different settings.
As described in~\cref{ap:implementation_swav}, we train each method on $64$ V100 16GB GPUs, with a batch size of $4096$, using mixed precision and \texttt{apex} optimized version of synchronized batch-normalization layers.
We report results with ResNet-$50$ for all methods.
In~\cref{fig:cost}, we report \OURS performance for different training lengths measured in hours based on our implementation.
We observe that after only $6$ hours of training, \OURS outperforms SimCLR trained for $1000$ epochs ($40$ hours based on our implementation) by a large margin.
If we train \OURS for longer, we see that the performance gap between the two methods increases even more.

\begin{table}[h]
  \caption{
\textbf{Computational cost.} We report time and GPU memory requirements based on our implementation for different models trained during $100$ epochs.
}
\vspace{.3em}
  \label{tab:cost}
\centering
\begin{tabular}{l c c c c}
    \toprule
	  Method & \texttt{multi-crop} & time / 100 epochs & peak memory / GPU \\
    \midrule
	SimCLR & $2 \times 224$ & 4h00 & 8.6G \\
	\OURS & $2 \times 224$ & 4h09 & 8.6G \\
	\OURS & $2 \times 160$ + $4 \times 96$ & 4h50 & 8.5G \\
	\OURS & $2 \times 224$ + $6 \times 96$ & 6h15 & 12.8G \\
    \bottomrule
  \end{tabular}
\end{table}

\subsubsection{Larger architectures}
In~\cref{tab:larger}, we show results when training \OURS on large architectures.
We observe that \OURS benefits from training on large architectures and plan to explore in this direction to furthermore boost self-supervised methods.
\begin{table}[h]
\centering
  \caption{
\textbf{Large architectures.}
Top-1 accuracy for linear models trained on frozen features from different self-supervised methods on large architectures.
}
\vspace{.3em}
\begin{tabular}{@{} l l c c @{}}
\toprule
Method      & Arch. & Param. & Top1  \\
\midrule
Supervised & EffNet-B7    & 66  & 84.4  \\
      \midrule
      Rotation~\cite{gidaris2018unsupervised} & RevNet50-4w     & 86  & 55.4 \\
      BigBiGAN~\cite{donahue2019large}        & RevNet50-4w     & 86  & 61.3 \\
      AMDIM~\cite{bachman2019learning}         & Custom-RN       & 626 & 68.1 \\
      CMC~\cite{tian2019contrastive}           & \resnetfifty-w2 & 188 & 68.4 \\
      MoCo~\cite{he2019momentum}               & \resnetfifty-w4 & 375 & 68.6 \\
      CPC v2~\cite{henaff2019data}             & R161            & 305 & 71.5 \\
      SimCLR~\cite{chen2020simple}             & \resnetfifty-w4 & 375 & 76.8 \\
      \midrule
      \OURS                             & \resnetfifty-w2 & 188 & \bf 77.3 \\
      \OURS                             & \resnetfifty-w4 & 375 & \bf 77.9 \\
      \OURS                             & \resnetfifty-w5 & 586 & \bf 78.5 \\
    \bottomrule
 \end{tabular}
  \label{tab:larger}
\end{table}
\paragraph{Implementation details for SwAV \resnetfifty-w2.}
The model is trained for $400$ epochs on $128$ GPUS (batch size $4096$).
We train the model with 2x224+4x96 (total of $6$ crops).
All other hyperparameters are the same as the ones described in appendix~\ref{ap:implementation_swav}.

\paragraph{Implementation details for SwAV \resnetfifty-w4.}
The model is trained for $400$ epochs on $64$ GPUS (batch size $2560$) with a queue of $2560$ samples starting from the beginning of training.
We train the model with 2x224+4x96 (total of $6$ crops).
All other hyperparameters are the same as the ones described in appendix~\ref{ap:implementation_swav}.

\paragraph{Implementation details for SwAV \resnetfifty-w5.}
The model is trained for $400$ epochs on $128$ GPUS (batch size $1536$) with a queue of $1536$ samples starting from the beginning of training.
We train the model with 2x224+4x96 (total of $6$ crops).
All other hyperparameters are the same as the ones described in appendix~\ref{ap:implementation_swav}.

\subsubsection{Transferring unsupervised features to downstream tasks}
In~\cref{tab:frozen_and_detect_ap}, we expand results from the main paper by providing numbers from previously and concurrently published self-supervised methods.
In the left panel of~\cref{tab:frozen_and_detect_ap}, we show performance after training a linear classifier on top of frozen representations on different datasets while on the right panel we evaluate the features by finetuning a ResNet-$50$ on object detection with Faster R-CNN~\cite{ren2015faster} and DETR~\cite{carion2020end}.
Overall, we observe on~\cref{tab:frozen_and_detect_ap} that \OURS is the first self-supervised method to outperform ImageNet supervised backbone on all the considered transfer tasks and datasets.
Other self-supervised learners are capable of surpassing the supervised counterpart but only for one type of transfer (object detection with finetuning for MoCo/PIRL for example).
We will release this model so other researchers might also benefit by replacing the ImageNet supervised network with our model.

\begin{table}[h]
\centering
    \caption{
\textbf{Transfer learning on downstream tasks.}
Comparison between features from ResNet-50 trained on ImageNet with \OURS or supervised learning.
We also report numbers from other self-supervised methods ($^\dagger$ for numbers from other methods run by us).
We consider two settings.
(1) Linear classification on top of frozen features. We report top-1 accuracy on all datasets except \VOCseven where we report mAP.
	(2) Object detection with finetuned features on \VOCseventwelve \texttt{trainval} using Faster R-CNN~\cite{ren2015faster} and on COCO~\cite{lin2014microsoft} using Mask R-CNN~\cite{he2017mask} or DETR~\cite{carion2020end}.
We report the most standard detection metrics for these datasets: $\text{AP}_{50}$ on \VOCseventwelve and $\text{AP}$ on COCO.
    }
    \label{tab:frozen_and_detect_ap}
    \vspace{.4em}
  \setlength{\tabcolsep}{3.7pt}
    \begin{tabular}{ @{} l ccc c ccc @{} }
      \toprule
       & \multicolumn{3}{c}{Linear Classification} &~~~~& \multicolumn{3}{c}{Object Detection}\\
\cmidrule{2-4}\cmidrule{6-8}
	    &\Places & \VOCseven & \iNat && \VOCseventwelve & COCO & COCO \\
	    & &  &  && \scriptsize(Faster R-CNN R50-C4) & \scriptsize(Mask R-CNN R50-FPN) & \scriptsize(DETR) \\
      \midrule
	    Supervised & $53.2^{\phantom{\dagger}}$  & $87.5^{\phantom{\dagger}}$ & $46.7^{\phantom{\dagger}}$ && $81.3^{\phantom{\dagger}}$ & $39.7^{\phantom{\dagger}}$ & $40.8^{\phantom{\dagger}}$ \\
      \midrule
	    RotNet~\cite{gidaris2020learning} & $45.0^{\phantom{\dagger}}$ & $64.6^{\phantom{\dagger}}$ & - && - & - \\
	    NPID++~\cite{misra2019self} & $46.4^{\phantom{\dagger}}$ & $76.6^{\phantom{\dagger}}$ & $32.4^{\phantom{\dagger}}$ && $79.1^{\phantom{\dagger}}$ & - \\
	    MoCo~\cite{he2019momentum} & $46.9^{\dagger}$ & $79.8^{\dagger}$ & $31.5^{\dagger}$ && $81.5^{\phantom{\dagger}}$ & - \\
	    PIRL~\cite{misra2019self} & $49.8^{\phantom{\dagger}}$ & $81.1^{\phantom{\dagger}}$ & $34.1^{\phantom{\dagger}}$ && $80.7^{\phantom{\dagger}}$ & - \\
	    PCL~\cite{li2020prototypical} & $49.8^{\phantom{\dagger}}$ & $84.0^{\phantom{\dagger}}$ & - && - & - \\
	    BoWNet~\cite{gidaris2020learning} & $51.1^{\phantom{\dagger}}$ & $79.3^{\phantom{\dagger}}$ & - && $81.3^{\phantom{\dagger}}$ & - \\
	    SimCLR~\cite{chen2020simple} & $53.3^{\dagger}$ & $86.4^{\dagger}$ & $36.2^{\dagger}$ && - & - \\
	    MoCov2~\cite{he2019momentum} & $52.9^{\dagger}$ & $87.1^{\dagger}$ & $38.9^{\dagger}$ && $82.5^{\phantom{\dagger}}$ & $39.8^{\phantom{\dagger}}$ & $42.0^{\dagger}$ \\
      \midrule
	    \OURS  & $\mathbf{56.7}^{\phantom{\dagger}}$ & $\mathbf{88.9}^{\phantom{\dagger}}$ & $\mathbf{48.6}^{\phantom{\dagger}}$ && $\mathbf{82.6}^{\phantom{\dagger}}$ & $\mathbf{41.6}^{\phantom{\dagger}}$ & $\mathbf{42.1}^{\phantom{\dagger}}$ \\
      \bottomrule
    \end{tabular}
\end{table}

\subsubsection{More detection metrics for object detection}
In~\cref{tab:cocomask}, ~\cref{tab:voc} and~\cref{tab:detr}, we evaluate the features by finetuning a ResNet-$50$ with different detectors: Mask R-CNN~\cite{he2017mask}, Faster R-CNN~\cite{ren2015faster} and DETR~\cite{carion2020end}.
We report more detection metrics compared to the table in the main paper~\cref{tab:frozen_and_detect_ap}.
We observe in~\cref{tab:cocomask}, \cref{tab:voc} and in~\cref{tab:detr} that \OURS outperforms the ImageNet supervised pretrained model on all the detection evaluation metrics.

Note that MoCov2 backbone performs particularly well on the object detection benchmark, and even outperform \OURS features for some detection metrics.
However, as shown in~\cref{tab:frozen_and_detect_ap}, this backbone is not competitive with the supervised features when evaluating on classification tasks without finetuning.

\begin{table}[h]
\centering
  \caption{
\textbf{Object detection and instance segmentation finetuned on COCO.}
Following the setup of \cite{he2019momentum}, we use Mask R-CNN detector with ResNet50-FPN trained on \texttt{train2017} with default $1 \times$ schedule and evaluated on \texttt{val2017}.
The metrics include bounding box AP (AP$^\text{b}$) and mask AP (AP$^\text{m}$).}
\vspace{.3em}
\begin{tabular}{l lll lll}
\toprule
	Method & AP$^{b}$ & AP$^{b}_{50}$ & AP$^{b}_{75}$ & AP$^{m}$ & AP$^{m}_{50}$ & AP$^{m}_{75}$ \\
\midrule
	Supervised  & 39.7 & 59.5 & 43.3 & 35.9 & 56.6 & 38.6 \\
\midrule
	MoCo-v2~\cite{chen2020improved} & 39.8 & 59.8 & 43.6 & 36.1 & 56.9 & 38.7 \\
\midrule
	\OURS  & \bf 41.6 & \bf 62.3 & \bf 45.5 & \bf 37.8 & \bf 59.6 & \bf 40.5 \\
\bottomrule
\end{tabular}
  \label{tab:cocomask}
\end{table}

\begin{table}[h]
\centering
  \caption{
\textbf{More detection metrics for object detection on \VOCseventwelve with finetuned features using Faster R-CNN~\cite{ren2015faster}.}
}
\vspace{.3em}
\begin{tabular}{l lll}
\toprule
Method & AP$^{all}$ & AP$^{50}$ & AP$^{75}$ \\
\midrule
Supervised                                   & 53.5 & 81.3 & 58.8 \\
Random                                            & 28.1 & 52.5 & 26.2 \\
\midrule
NPID++~\cite{misra2019self} & 52.3 & 79.1 & 56.9 \\
PIRL~\cite{misra2019self} & 54.0 & 80.7 & 59.7 \\
BoWNet~\cite{gidaris2020learning} & 55.8 & 81.3 & 61.1 \\
MoCov1~\cite{he2019momentum} & 55.9 & 81.5 & 62.6 \\
MoCov2~\cite{chen2020improved} & \bf 57.4 & 82.5 & \bf 64.0 \\
\midrule
\OURS  & 56.1 & \bf 82.6 & 62.7 \\
\bottomrule
\end{tabular}
  \label{tab:voc}
\end{table}

\begin{table}[h]
\centering
  \caption{
\textbf{More detection metrics for object detection on COCO with finetuned features using DETR~\cite{carion2020end}.}
}
\vspace{.3em}
\begin{tabular}{ll c c c c c c}
    \toprule
	  Method && $\text{AP}$ & $\text{AP}_{50}$ & $\text{AP}_{75}$ & $\text{AP}_S$ & $\text{AP}_M$ & $\text{AP}_L$ \\
    \midrule
	\ImNet labels && 40.8 & 61.2 & 42.9 & 20.1 & 44.5 & 60.3 \\
    \midrule
	MoCo-v2 && 42.0 & 62.7 & 44.4 & \bf 20.8 & 45.6 & \bf 60.9 \\
	\OURS && \bf 42.1 & \bf 63.1 & \bf 44.5 & 19.7 & \bf 46.3 & \bf 60.9 \\
    \bottomrule
  \end{tabular}
  \label{tab:detr}
\end{table}

\subsubsection{Low-Shot learning on ImageNet for \OURS pretrained on Instagram data}
We now test whether \OURS pretrained on Instagram data can serve as a pretraining method for low-shot learning on ImageNet.
We report in~\cref{tab:detr} results when finetuning Instagram \OURS features with only few labels per ImageNet category.
We observe that using pretrained features from Instagram improves considerably the performance compared to training from scratch.
\begin{table}[h]
  \caption{
\textbf{Low-shot learning on ImageNet.} Top-1 and top-5 accuracies when training with $13$ or $128$ examples per category.
}
\vspace{.3em}
  \label{tab:semisup}
\centering
\begin{tabular}{ll c c c c}
    \toprule
	  \# examples per class & \multicolumn{2}{c}{13} & \multicolumn{2}{c}{128} \\
	  & top1 & top5 & top1 & top5 \\
    \midrule
	No pretraining & 25.4 & 48.4 & 56.4 & 80.4 \\
    \midrule
	\OURS IG-1B & \bf 38.2  & \bf 67.1  & \bf 64.7 & \bf 87.2 \\
    \bottomrule
  \end{tabular}
\end{table}

\subsubsection{Image classification with KNN classifiers on \ImNet}
Following previous work protocols~\cite{wu2018unsupervised,zhuang2019local}, we evaluate the quality of our unsupervised features with K-nearest neighbor (KNN) classifiers on ImageNet.
We get features from the computed network outputs for center crops of training and test images.
We report results with $20$ and $200$ NN in~\cref{tab:knn}.
We outperform the current state-of-the-art of this evaluation.
Interestingly we also observe that using fewer NN actually boosts the performance of our model.
\begin{table}[h]
  \caption{
\textbf{KNN classifiers on ImageNet.} We report top-1 accuracy with $20$ and $200$ nearest neighbors.
}
\vspace{.3em}
  \label{tab:knn}
\centering
\begin{tabular}{ll c c}
    \toprule
	  Method && 20-NN & 200-NN \\
    \midrule
	NPID~\cite{wu2018unsupervised} && - & 46.5 \\
	LA~\cite{zhuang2019local} && - & 49.4 \\
	PCL~\cite{li2020prototypical} && 54.5 & - \\
    \midrule
	\OURS && \bf 65.7 & \bf 62.7 \\
    \bottomrule
  \end{tabular}
\end{table}

\subsection{Ablation Studies on Clustering}
\label{ap:ablation}

\subsubsection{Number of prototypes}
In~\cref{tab:nmbproto}, we evaluate the influence of the number of prototypes used in \OURS.
We train ResNet-50 with \OURS for $400$ epochs with $2 \times 160 + 4 \times 96$ crops (ablation study setting) and evaluate the performance by training a linear classifier on top of frozen final representations.
We observe in~\cref{tab:nmbproto} that varying the number of prototypes by an order of magnitude (3k-100k) does not affect much the performance (at most $0.3$ on ImageNet).
This suggests that the number of prototypes has little influence as long as there are ``enough''.
Throughout the paper, we train \OURS with $3000$ prototypes.
We find that using more prototypes increases the computational time both in the Sinkhorn algorithm and during back-propagation for an overall negligible gain in performance.

\begin{table}[h]
\centering
  \caption{
\textbf{Impact of number of prototypes.} Top-1 ImageNet accuracy for linear models trained on frozen features.
}
\vspace{.3em}
\begin{tabular}{l c c c c c c}
    \toprule
	  Number of prototypes & 300 & 1000 & 3000 & 10000 & 30000 & 100000 \\
    \midrule
	Top-1 & 72.8 & 73.6 & 73.9 & 74.1 & 73.8 & 73.8 \\
    \bottomrule
  \end{tabular}
  \label{tab:nmbproto}
\end{table}

\subsubsection{Learning the prototypes}
We investigate the impact of learning the prototypes compared to using fixed random prototypes.
Assigning features to fixed random targets has been explored in NAT~\cite{bojanowski2017unsupervised}.
However, unlike \OURS, NAT uses a target per instance in the dataset, the assignment is hard and performed with Hungarian algorithm.
In~\cref{tab:learned_soft} (left), we observe that learning prototypes improves \OURS from $73.1$ to $73.9$ which shows the effect of adapting the prototypes to the dataset distribution.

Overall, these results suggest that our framework learns from a different signal from "offline" approaches that attribute a pseudo-label to each instance while considering the full dataset and then predict these labels (like DeepCluster~\cite{caron2018deep} for example). 
Indeed, the prototypes in \OURS are not strongly encouraged to be categorical and random fixed prototypes work almost as well.
Rather, they help contrasting different image views without relying on pairwise comparison with many negatives samples.
This might explain why the number of prototypes does not impact the performance significantly.

\begin{table}[h]
  \caption{
\textbf{Ablation studies on clustering.} Top-1 ImageNet accuracy for linear models trained on frozen features.
\textbf{(left)} Impact of learning the prototypes.
\textbf{(right)} Hard versus soft assignments.
}
\label{tab:learned_soft}
\vspace{.3em}
\begin{minipage}{0.5\linewidth}
\centering
\begin{tabular}{l c c}
    \toprule
	  Prototypes & Learned & Fixed \\
    \midrule
	Top-1 & 73.9 & 73.1 \\
    \bottomrule
  \end{tabular}
\end{minipage}
\begin{minipage}{0.5\linewidth}
\begin{tabular}{l c c}
    \toprule
	  Assignment & Soft & Hard \\
    \midrule
	Top-1 & 73.9 & 73.3 \\
    \bottomrule
  \end{tabular}
\end{minipage}
\end{table}

\begin{figure}[b]
\begin{minipage}{0.48\linewidth}
\centering
\includegraphics[width=\linewidth]{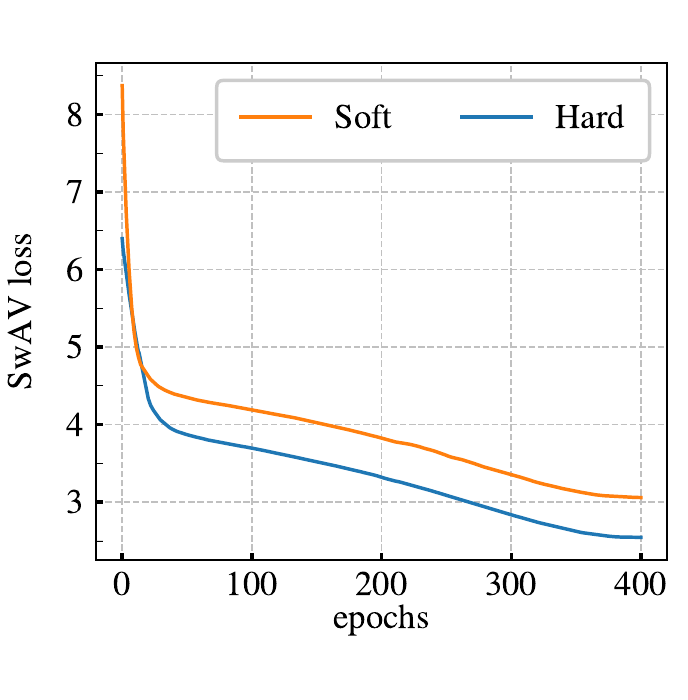}
\end{minipage}
~~~~~\begin{minipage}{0.48\linewidth}
\caption{\textbf{Hard versus soft assignments.}
We report the training loss for \OURS models trained with either soft or hard assignments.
The models are trained during $400$ epochs with $2 \times 160 + 4 \times 96$ crops.
}
\label{fig:losssoft}
\end{minipage}
\end{figure}

\subsubsection{Hard versus soft assignments}
In~\cref{tab:learned_soft} (right), we evaluate the impact of using hard assignment instead of the default soft assignment in \OURS.
We train the models during $400$ epochs with $2 \times 160 + 4 \times 96$ crops (ablation study setting) and evaluate the performance by training a linear classifier on top of frozen final representations.
We also report the training losses in~\cref{fig:losssoft}.
We observe that using the hard assignments performs worse than using the soft assignments.
An explanation is that the rounding needed to obtain discrete codes is a more aggressive optimization step than gradient updates.
While it makes the model converge rapidly (see \cref{fig:losssoft}), it leads to a worse solution.

\begin{table}[b]
\centering
  \caption{
\textbf{Impact of the number of iterations in Sinkhorn algorithm.} Top-1 ImageNet accuracy for linear models trained on frozen features.
}
\vspace{.3em}
\begin{tabular}{l c c c c}
    \toprule
	Sinkhorn iterations & 1 & 3 & 10 & 30 \\
    \midrule
	Top-1 & \textit{fail} & 73.9 & 73.8 & 73.7 \\
    \bottomrule
  \end{tabular}
  \label{tab:iters}
\end{table}

\subsubsection{Impact of the number of iterations in Sinkhorn algorithm}
In~\cref{tab:iters}, we investigate the impact of the number of normalization steps performed during Sinkhorn-Knopp algorithm~\cite{cuturi2013sinkhorn} on the performance of \OURS.
We observe that using only $3$ iterations is enough for the model to converge.
When performing less iterations, the loss fails to converge.
We observe that using more iterations slightly alters the transfer performance of the model.
We conjecture that it is for the same reason that rounding codes to discrete values deteriorate the quality of our model by converging too rapidly.

\subsection{Details on Clustering-Based methods: DeepCluster-v2 and SeLa-v2}
\label{ap:clustering}
In this section, we provide details on our improved implementation of clustering-based approaches DeepCluster-v2 and SeLa-v2 compared to their corresponding original publications~\cite{caron2018deep,asano2019self}.
These two methods follow the same pipeline: they alternate between pseudo-labels generation (``assignment phase'') and training the network with a classification loss supervised by these pseudo-labels (``training phase'').

\subsubsection{Training phase}
During the training phase, both methods minimize the multinomial logistic loss of the pseudo-labels $\mathbf{q}$ classification problem:
\begin{equation}
  \ell(\mathbf{z}, \mathbf{c}, \mathbf{q}) = - \sum_{k} \mathbf{q}^{(k)} \log \mathbf{p}^{(k)}, \quad \text{where} \quad \mathbf{p}^{(k)} = \frac{ \exp \left ( \frac{1}{\tau} \mathbf{z}^\top \mathbf{c}_k \right ) }{\sum_{k'} \exp \left ( \frac{1}{\tau} \mathbf{z}^\top \mathbf{c}_{k'} \right ) }.
  \label{eq:ap_assignloss}
\end{equation}
The pseudo-labels are kept fixed during training and updated for the entire dataset once per epoch during the assignment phase.

\paragraph{Training phase in DeepCluster-v2.}
In the original DeepCluster work, both the classification head $\mathbf{c}$ and the convnet weights are trained to classify the images into their corresponding pseudo-label between two assignments.
Intuitively, this classification head is optimized to represent prototypes for the different pseudo-classes.
However, since there is no mapping between two consecutive assignments: the classification head learned during an assignment becomes irrelevant for the following one.
Thus, this classification head needs to be re-set at each new assignment which considerably disrupts the convnet training.
For this reason, we propose to simply use for classification head $\mathbf{c}$ the centroids given by k-means clustering (Eq.~\ref{eq:kmeans}).
Overall, during training, DeepCluster-v2 optimizes the following problem with mini-batch SGD:
\begin{equation}
\min_{\mathbf{z}} \ell(\mathbf{z}, \mathbf{c}, \mathbf{q}).
\end{equation}

\paragraph{Training phase in SeLa-v2.}
In SeLa work, the prototypes $\mathbf{c}$ are learned with stochastic gradient descend during the training phase.
Overall, during training, SeLa-v2 optimizes the following problem:
\begin{equation}
\min_{\mathbf{z},\mathbf{c}} \ell(\mathbf{z}, \mathbf{c}, \mathbf{q}).
\end{equation}

\subsubsection{Assignment phase}
The purpose of the assignment phase is to provide assignments $\mathbf{q}$ for each instance of the dataset.
For both methods, this implies having access to feature representations $\mathbf{z}$ for the entire dataset.
Both original works~\cite{caron2018deep,asano2019self} perform regularly a pass forward on the whole dataset to get these features.
Using the original implementation, if assignments are updated at each epoch, then the assignment phase represents one third of the total training time.
Therefore, in order to speed up training, we choose to use the features computed during the previous epoch instead of dedicating pass forwards to the assignments.
This is similar to the memory bank introduced by Wu~\etal~\cite{wu2018unsupervised}, without momentum.

\paragraph{Assignment phase in DeepCluster-v2.}
DeepCluster-v2 uses spherical k-means to get pseudo-labels.
In particular, pseudo-labels $\mathbf{q}$ are obtained by minimizing the following problem:
\begin{equation}
\label{eq:kmeans}
  \min_{\mathbf{C} \in \mathbb{R}^{d\times K}}
  \frac{1}{N}
  \sum_{n=1}^N
  \min_{\mathbf{q}}
	- \mathbf{z}_n^\top \mathbf{C} \mathbf{q},
\end{equation}
where $\mathbf{z}_n$ and the columns of $\mathbf{C}$ are normalized.
The original work DeepCluster uses tricks such as cluster re-assignments and balanced batch sampling to avoid trivial solutions but we found these unnecessary, and did not observe collapsing during our trainings.
As noted by Asano~\etal, this is due to the fact that assignment and training are well separated phases.

\paragraph{Assignment phase in SeLa-v2.}
Unlike DeepCluster, SeLa uses the same loss during training and assignment phases.
In particular, we use Sinkhorn-Knopp algorithm to optimize the following assignment problem (see details and derivations in the original SeLa paper~\cite{asano2019self}):
\begin{equation}
\min_{\mathbf{q}} \ell(\mathbf{z}, \mathbf{c}, \mathbf{q}).
\end{equation}

\paragraph{Implementation details}
We use the same hyperparameters as \OURS to train SeLa-v2 and DeepCluster-v2: these are described in~\cref{ap:implementation}.
Asano~\etal~\cite{asano2019self} have shown that multi-clustering boosts performance of clustering-based approaches, and so we use $3$ sets of $3000$ prototypes $\mathbf{c}$ when training SeLa-v2 and DeepCluster-v2.
Note that unlike online methods (like \OURS, SimCLR and MoCo), the clustering approaches SeLa-v2 and DeepCluster-v2 can be implemented with only a single crop per image per batch.
The major limitation of SeLa-v2 and DeepCluster-v2 is that these methods are not online and therefore scaling them to very large scale dataset is not posible without major adjustments.

\end{document}